\documentclass[10pt,twocolumn,letterpaper]{article}

\usepackage{cvpr}              % To produce the CAMERA-READY version
\usepackage{algorithm}
\usepackage{float}
\usepackage{algorithmic}
\usepackage{makecell}
\usepackage{multirow}
\usepackage{pifont}
\usepackage{adjustbox}
\usepackage{mathtools}
\usepackage{graphicx}
\usepackage{xcolor}
\usepackage{nicefrac}

\usepackage[accsupp]{axessibility}  % Improves PDF readability for those with disabilities.
\definecolor{cvprblue}{rgb}{0.21,0.49,0.74}
\usepackage[pagebackref,breaklinks,colorlinks,allcolors=cvprblue]{hyperref}
\newcommand{\hut}[1]{\textcolor{black}{#1}}
\newcommand{\zjn}[1]{\textcolor{black}{#1}} % Jiangning Zhang
\newcommand{\yr}[1]{\textcolor{black}{#1}}
\newcommand{\yrn}[1]{\textcolor{black}{#1}}

\newcommand{\hutnew}[1]{\textcolor{black}{#1}}

\def\paperID{3025} % *** Enter the Paper ID here
\def\confName{CVPR}
\def\confYear{2025}

\title{UltraGen: High-Resolution Video Generation with Hierarchical Attention}

\author{Teng Hu$^{1\ast}$
\quad Jiangning Zhang$^{2}$\thanks{Equal contribution.}
\quad Zihan Su$^{1}$
\quad Ran Yi$^{1}$\thanks{Corresponding author.}\\
\normalsize $^1$Shanghai Jiao Tong University \quad $^2$Zhejiang University\\
\normalsize $\{$hu-teng, ranyi$\}$@sjtu.edu.cn \quad $186368$@zju.edu.cn\\
{\tt\small Project page: \href{https://sjtuplayer.github.io/projects/UltraGen}{\textcolor{magenta}{https://sjtuplayer.github.io/projects/UltraGen}}}
}

\begin{document}
% \maketitle

\twocolumn[{%
\maketitle
\begin{figure}[H]
\hsize=\textwidth % cvpr 需要
\centering
\vspace{-0.2in}
\includegraphics[width=\textwidth]{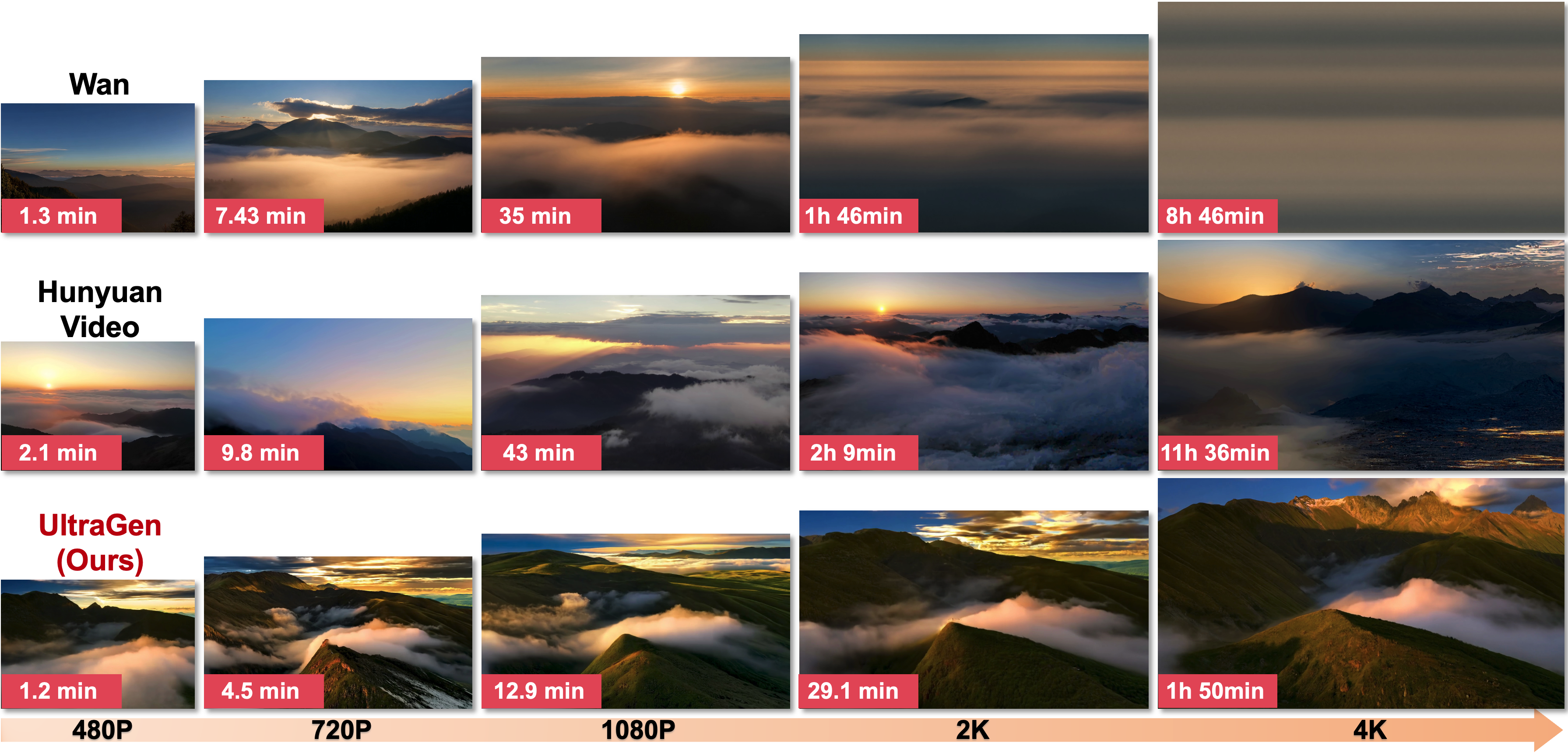}
\caption{
  Typical video generation models exhibit significant \textit{\begin{adjustbox}{valign=c}{\includegraphics[width=0.017\linewidth]{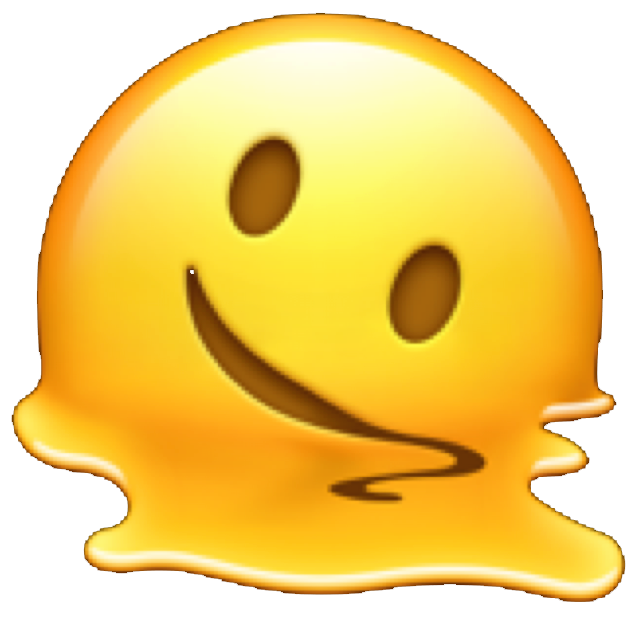}}\end{adjustbox}quality degradation} and \textit{\begin{adjustbox}{valign=c}{\includegraphics[width=0.017\linewidth]{figures/e1.png}}\end{adjustbox}increased processing time} with higher resolutions, whereas our UltraGen delivers \textit{\begin{adjustbox}{valign=c}{\includegraphics[width=0.017\linewidth]{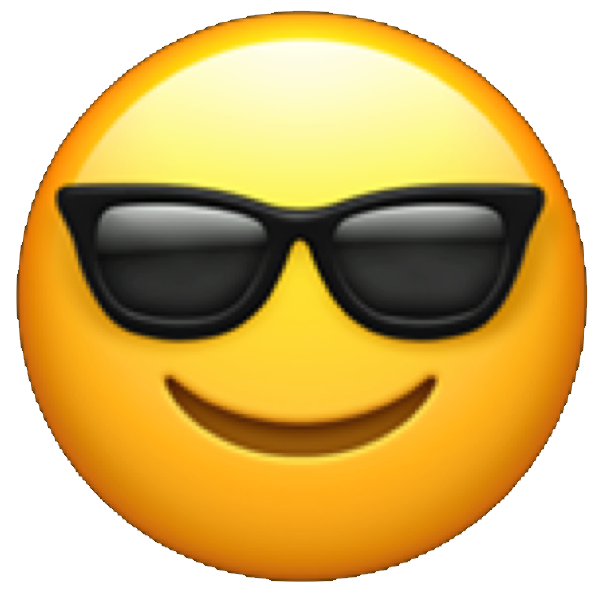}}\end{adjustbox}superior video quality} at resolutions beyond 2K while achieving \textit{\begin{adjustbox}{valign=c}{\includegraphics[width=0.017\linewidth]{figures/e2.png}}\end{adjustbox}4.78$\times$ speedup} compared to the popular Wan-T2V-1.3B baseline~\cite{wang2025wan} (81 frames, 4$\times$H20 GPUs). Enlarge for better visual effects.
  }
  \label{fig:teaser}
\end{figure}
}]

% \begin{figure*}[H]
% \hsize=\textwidth % cvpr 需要
% \centering
%   \includegraphics[width=\textwidth]{figures/teaser1.png}
%   \vspace{-0.1in}
%   \caption{
%   Typical video generation models exhibit significant \textit{\begin{adjustbox}{valign=c}{\includegraphics[width=0.017\linewidth]{figures/e1.png}}\end{adjustbox}quality degradation} and \textit{\begin{adjustbox}{valign=c}{\includegraphics[width=0.017\linewidth]{figures/e1.png}}\end{adjustbox}increased processing time} with higher resolutions, whereas our UltraGen delivers \textit{\begin{adjustbox}{valign=c}{\includegraphics[width=0.017\linewidth]{figures/e2.png}}\end{adjustbox}superior video quality} at resolutions beyond 2K while achieving \textit{\begin{adjustbox}{valign=c}{\includegraphics[width=0.017\linewidth]{figures/e2.png}}\end{adjustbox}4.78$\times$ speedup} compared to the popular Wan-T2V-1.3B baseline~\cite{wang2025wan} (81 frames, 4$\times$H20 GPUs). Enlarge for better visual effects.
% \end{figure*}

\begin{abstract}
Recent advances in video generation have made it possible to produce visually compelling videos, with wide-ranging applications in content creation, entertainment, and virtual reality.
However, most existing diffusion transformer based video generation models are limited to low-resolution outputs ($\leq$720P) due to the quadratic computational complexity of the attention mechanism with respect to
the output width and height.
This computational bottleneck makes native high-resolution video generation (1080P/2K/4K) impractical for both training and inference.
To address this challenge, we present \textbf{UltraGen}, a novel video generation framework that enables \textbf{\textit{i) efficient}} and \textbf{\textit{ii) end-to-end native high-resolution}} video synthesis.
Specifically, UltraGen features a hierarchical dual-branch attention architecture based on global-local attention decomposition, which decouples full attention into a local attention branch for high-fidelity regional content and a global attention branch for overall semantic consistency.
We further propose a spatially compressed global modeling strategy to efficiently learn global dependencies, and a hierarchical cross-window local attention mechanism to reduce computational costs while enhancing information flow across different local windows.
Extensive experiments demonstrate that UltraGen can effectively scale pre-trained low-resolution video models to 1080P and even 4K resolution for the first time, outperforming existing state-of-the-art methods and super-resolution based two-stage pipelines in both qualitative and quantitative evaluations. 
\end{abstract}    
\section{Introduction}
The field of video generation~\cite{huang2024make,villegas2022phenaki,singer2022make,adavideorag} has undergone rapid development in recent years, unlocking a diverse array of downstream applications, including video customization~\cite{hu2025polyvivid,hu2025hunyuancustom,liu2025phantom}, video editing~\cite{liang2025omniv2v,ivebench,jiang2025vace}, and video motion control~\cite{hu2024animate,hu2024motionmaster,hu2025high}.
With the emergence of powerful diffusion-based generative models~\cite{ho2020ddpm}, the quality, coherence, and diversity of generated videos have significantly improved, narrowing the gap between synthetic and real-world content. \hut{Based on diffusion transformers~\cite{dit},} state-of-the-art models such as Wan~\cite{wang2025wan} and HunyuanVideo~\cite{hunyuanvideo} have demonstrated impressive capabilities in synthesizing temporally consistent and semantically rich videos, making remarkable progress in high-quality video generation.

Despite these advancements, current video generation models still suffer from a critical limitation: restricted resolution. 
% Even the most advanced models to date are typically constrained to generating videos at or below 720P resolution. 
\hut{Since the advanced video generation models~\cite{wang2025wan,hunyuanvideo} are based on diffusion transformers~\cite{dit}, they inherently suffer from} the quadratic computational complexity of the full-attention mechanism with respect to the spatiotemporal size of the input, \textit{i.e.,} $\mathcal{O}((T \cdot H \cdot W)^2)$, where $T$, $H$, and $W$ denote the temporal length, height, and width of the video, respectively. \yr{For instance,} doubling the \yr{width and height will} result in a 16-fold increase in computational cost, making high-resolution video generation prohibitively expensive for both training and inference. To mitigate this, existing approaches~\cite{he2022latent,blattmann2023align,singer2022make} often resort to a two-stage pipeline that first generates low-resolution videos and subsequently applies video super-resolution models. However, this paradigm merely enhances visual clarity and fails to introduce \yr{enough} visual details, leading to the synthesis of pseudo high-resolution content with limited authenticity and richness.

To address these challenges, we propose \textbf{UltraGen}, a hierarchical attention-based framework for native high-resolution video generation. \textit{UltraGen} offers an efficient and scalable solution that transforms pre-trained low-resolution video diffusion models into end-to-end high-resolution generators with significantly reduced computational overhead. Concretely, we \yr{propose} a dual-branch video generation architecture that decouples the full attention mechanism into \textit{local} and \textit{global \yr{attention}} branches. The local attention branch focuses on generating fine-grained %high-resolution 
content within individual \yr{local spatial windows}, while the global \yr{attention} branch captures holistic video semantics and ensures coherence across different local \yr{windows}. 
To efficiently model global dependencies without incurring prohibitive costs, we design a \textbf{spatially compressed global modeling module} that compresses spatial information via frame-wise convolutions before applying attention, \yr{so that the self-attention is conducted at a smaller spatial size,} followed by 3D convolutions to restore spatial fidelity and enhance temporal continuity.
Furthermore, to ensure effective information flow across \yr{different} local \yr{windows}, we propose a \textbf{hierarchical cross-window local attention mechanism}. By %introducing shared regions between adjacent local attention modules, 
\yr{partitioning the local windows of adjacent layers differently and creating intersections between them,}
our model enables seamless interaction and consistency across spatial local \yr{windows}, further improving the video generation quality.

We conduct extensive experiments by extending the Wan-1.3B model to support native 1080P and 4K video generation, which is the first to achieve native high-quality 4K synthesis in the field. Comparisons against state-of-the-art models, including Wan and Hunyuan Video, as well as two-stage pipelines (low-resolution generation + super resolution), demonstrate that \textit{UltraGen} significantly outperforms existing methods both qualitatively and quantitatively, validating the effectiveness and scalability of our approach.

\begin{itemize}
    \item We propose \textbf{UltraGen}, a novel high-resolution video generation framework based on global-local attention decomposition, which enables scalable extension of low-resolution pre-trained video diffusion models to support 1080P and 4K resolution in an end-to-end manner.
    
    \item We design a \textbf{Spatially Compressed Global Attention Mechanism} that significantly reduces computation cost of global context modeling. By compressing spatial information via frame-wise convolution, \yr{conducting self-attention at a smaller spatial size}, and decoding through 3D convolution, our method \yr{efficiently} captures holistic semantics while keeping temporal coherence.
    
    \item We introduce a \textbf{Hierarchical Cross-window Local Attention Mechanism} that facilitates efficient interaction among local regions. By allowing intersecting regions between attention windows of adjacent layers, it ensures smooth content transitions and enhances local detail.
    
    \item UltraGen is the first model to achieve \textbf{native high-quality 4K video generation}. Extensive experiments demonstrate its superior ability in HD video generation.
    % that our model outperforms state-of-the-art baselines, including Wan, Hunyuan Video, and two-stage low-resolution + super-resolution pipelines, across multiple metrics and resolutions.
\end{itemize}

\section{Related Work}

\subsection{%Fundamental 
Video Generation Foundation Models} 

The advent of diffusion models~\cite{ho2020ddpm} has greatly advanced video generation. Early methods~\cite{guo2023animatediff,blattmann2023svd} typically extend text-to-image diffusion models~\cite{rombach2022ldm} by adding temporal modules to capture frame dynamics. While somewhat effective, these approaches often separate spatial and temporal modeling, limiting their ability to capture holistic spatiotemporal dependencies and resulting in less coherent videos.
With DiT~\cite{flux2024}, transformer-based architectures have become the leading paradigm in video generation~\cite{yang2024cogvideox,zheng2024opensora}. These models treat videos as spatiotemporal volumes, flattening them into 1D token sequences across time, height, and width. Full self-attention is then used to jointly model spatial and temporal relationships, leading to notable improvements in temporal consistency and spatial detail.
Recent work has further advanced video generation by leveraging large transformer backbones and massive video datasets. Notably, models like Wan~\cite{wang2025wan} and HunyuanVideo~\cite{hunyuanvideo} show that scaling up model size and data significantly enhances video quality and diversity. These models achieve impressive text-to-video synthesis, producing videos with rich content and improved temporal consistency. However, \textit{due to the quadratic complexity of self-attention, they remain limited to relatively low resolutions (\textit{e.g.}, 720P), and scaling to higher resolutions is still a major challenge.}

\begin{figure*}[t]
    \centering
    \includegraphics[width=0.9\textwidth]{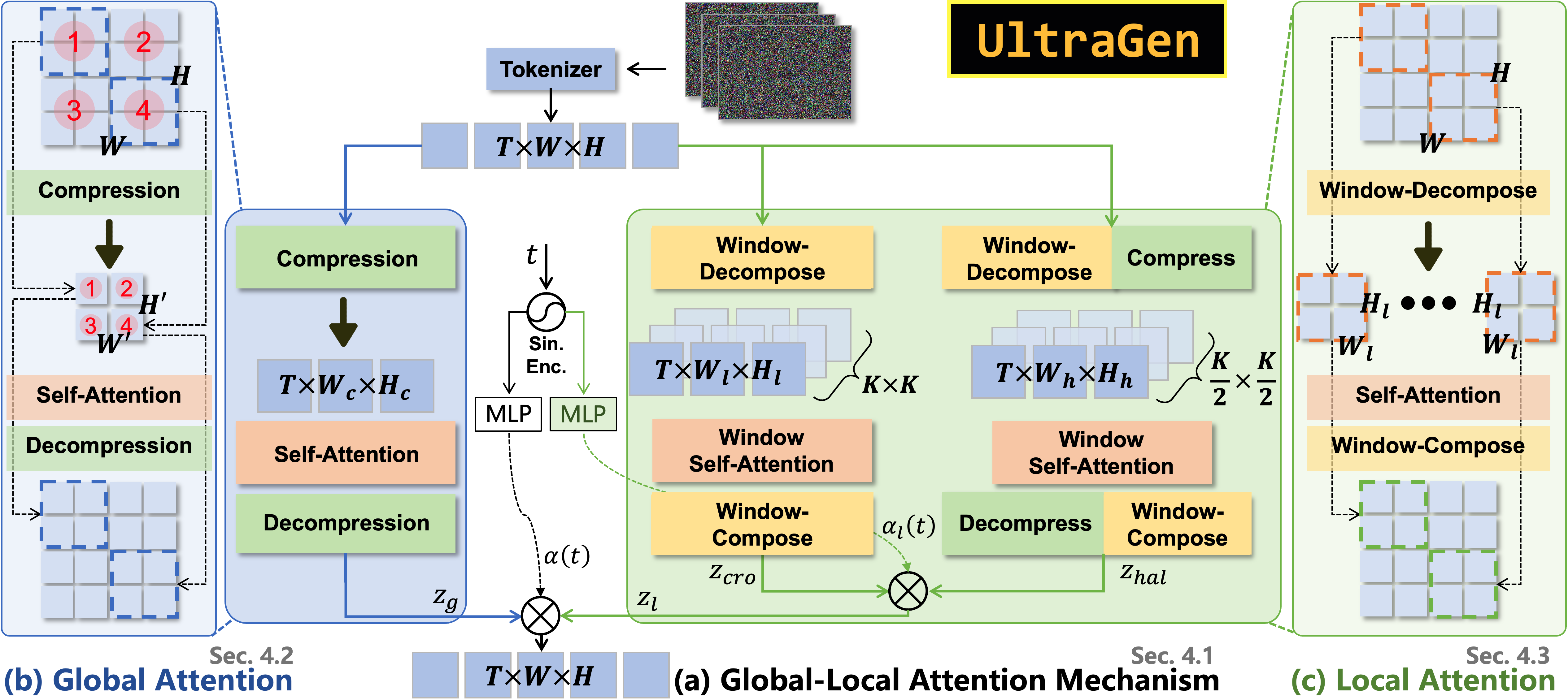}
    \caption{\zjn{\textbf{Overview of our UltraGen} that decomposes the full-attention into a global \yr{attention} branch (Sec.~\ref{sec:global}) for overall semantic consistency and a local \yr{attention} branch (Sec.~\ref{sec:local}) for high-fidelity regional content, boosting high-efficiency and high-resolution video generation.}}
    \label{fig:framework}
    \vspace{-0.1in}
\end{figure*}

\subsection{High-resolution Video Generation}

To enable high-resolution generation, some existing methods such as Wan~\cite{wang2025wan} and HunyuanVideo~\cite{hunyuanvideo} train their models on videos of various resolutions, allowing them to scale to arbitrary output sizes. However, when generating videos at resolutions beyond 2K, these approaches often produce blurry results, as illustrated in Fig.~\ref{fig:teaser}.
In contrast to directly modeling high-resolution generation, other methods~\cite{he2022latent,blattmann2023align,singer2022make,Ho2022ImagenVH,wang2024lavie}, such as Align-Your-Latents~\cite{blattmann2023align}, adopt a two-stage process: they first generate low-resolution videos and then apply super-resolution~\cite{du2024ld,zhou2024upscale,zhang2024realviformer} to upscale the output. However, super-resolution primarily improves visual sharpness without introducing sufficient new details, resulting in pseudo high-resolution content that lacks authenticity and richness.
\hutnew{Some recent works~\cite{wang2025lingen,dalal2025oneminute} have made progress in long video generation by leveraging linear attention mechanisms~\cite{gu2023mamba} or test-time training~\cite{zhang2025TTT}; however, they have paid limited attention to scaling up the spatial resolution of videos. To address these challenges, we investigate native high-definition (HD) video generation, aiming to overcome the high computational costs while producing high-quality HD videos.}

\section{Preliminaries}
\label{sec:preliminary}

\textbf{\yr{DiT-based Video Generation.}} Most state-of-the-art \yr{Diffusion Transformer (DiT) based} video generation models (\textit{e.g.}, Wan~\cite{wang2025wan} and HunyuanVideo~\cite{hunyuanvideo}) adopt a full-attention-based framework, which builds upon the Transformer architecture to model spatiotemporal dependencies in video sequences. Typically, a 3D variational autoencoder (3D-VAE) is first used to encode an input video into a latent representation of shape $\yr{D} \times T \times H \times W$, where $\yr{D}$ denotes the %number of channels, 
\yr{hidden dimension,} $T$, $H$, and $W$ represent the temporal frames, height, and width, respectively. 
% In practice, to reduce the computational burden of subsequent attention layers, the spatial and temporal resolutions are typically downsampled by factors of $8$ and $4$, respectively.
This downsampling strategy effectively reduces the sequence length and makes training tractable for medium-sized videos. 
% To adapt this high-dimensional tensor for transformer-based processing, the latent video is reshaped into a 1D token sequence via a patchify module. Specifically, the spatiotemporal tensor is flattened into a sequence of $N = T \cdot H \cdot W$ tokens, each of dimension $\yr{D}$. 
\hutnew{Then, the video latents are reshaped into a 1D token sequence with sequence length $N = T \times H \times W$ via a patchify module.}

\hutnew{Once the token sequence is obtained, video generation models apply full self-attention mechanisms across the entire sequence. For a sequence of $N$ tokens, the self-attention module computes an $N \times N$ attention map, which scales quadratically with the sequence length.}

\hutnew{The computational complexity of self-attention is $\mathcal{O}(N^2 \cdot D)$, 
which becomes prohibitively expensive as the video resolution increases. For instance, doubling the height and width of the video leads to a four-fold increase in the number of tokens and a sixteen-fold increase in the size of the attention map. This quadratic scaling severely limits the feasibility of generating high-resolution videos (\textit{e.g.}, 1080P and even 4K) using existing full-attention architectures in terms of training and inference costs.}

\section{UltraGen: Born for HD Video Generation}
\label{sec:method}

\subsection{Time-Aware Global-Local Attention } \label{sec:global_local}

As discussed in Sec.~\ref{sec:preliminary}, in DiT-based video generation, the computational complexity of full attention is $\mathcal{O}((TWH)^2 \times D)$, which grows quadratically with the spatial size $(W \times H)$ of the generated video. To address this, we restrict attention to a fixed local region by introducing an attention window of size $(W_0, H_0)$. This ensures that, regardless of the overall spatial dimensions, attention is computed only within each $(W_0, H_0)$ window. By applying this \textit{local attention mechanism} to cover the entire frame, the total computational cost increases only linearly with the number of windows, rather than quadratically with frame size. Thus, the overall complexity is reduced to $\mathcal{O}((TW_0H_0)^2 \times D)$ up to a constant factor, effectively avoiding quadratic scaling. However, relying solely on local attention ignores dependencies across windows, potentially leading to isolated or inconsistent content. To address this, we introduce a \textit{global attention mechanism} that connects all local windows, enabling the model to capture long-range dependencies and maintain semantic consistency across the frame, thereby supporting high-resolution video generation with coherent semantics.

\hutnew{Therefore, we propose a novel \textbf{global-local attention mechanism} that decomposes the original full attention module into two complementary components: \textit{global attention} and \textit{local attention}. Specifically, the local attention module partitions the video sequence into multiple independent sub-regions and applies attention within each region separately, significantly reducing the overall computational cost. In parallel, the global attention module models the interactions across different local regions, injecting holistic spatiotemporal information into each local branch. This hierarchical design enables efficient and scalable attention modeling while preserving both local detail and global coherence.}

\noindent\textbf{Local Attention Mechanism.} 
For a video latent representation $z_v\in\mathcal{R}^{B\times (T\cdot W\cdot H)\times D}$, we aim to reduce the computational burden of self-attention by introducing a \textbf{local attention mechanism} that approximates full self-attention with lower computational cost.

\hutnew{We partition the video latent $z_v$ along the spatial dimensions ($H$ and $W$) into $m$ non-overlapping, equally sized local windows, each with dimensions ${B \times (T\cdot W_0
\cdot H_0)\times D}$. For each local window, self-attention is applied independently, and the results are aggregated along the spatial dimensions to produce the final local attention output with the original resolution:}
\begin{equation}
\begin{aligned}
    &\{z^{(i)}_l\}_{i=1}^m = Partition(z_v), \\
    &z'^{(i)}_l = Self\text{-}Attention(z^{(i)}_l), \\
    &z_l = Aggregate(\{z'^{(i)}_l\}_{i=1}^m),
\end{aligned}
\end{equation}
\hutnew{where $Partition(\cdot)$ divides $z_v$ into $m$ local windows, $Self\text{-}Attention(\cdot)$ is applied within each window, and $Aggregate(\cdot)$ concatenates the outputs along the spatial dimensions to reconstruct the local attention result $z_l \in \mathbb{R}^{B \times T \times H \times W \times D}$ (detailed designs are in Sec.~\ref{sec:local}).}

\noindent\textbf{Global Attention Mechanism.} \hutnew{Local attention reduces computational cost but limits focus to individual windows, potentially causing semantic inconsistencies. For example, a prompt describing "a dog" might lead to multiple independent versions across windows.}

\hutnew{To address this, we introduce a \textbf{global attention module} to capture long-range dependencies and ensure semantic consistency. We compress the spatial information of the video latent $z_v$ into a lower-resolution $z_g \in \mathbb{R}^{B \times (T \cdot H_g \cdot W_g) \times D}$ using a convolution module, apply global self-attention at this reduced size, and decompress the result to the original resolution:}
\begin{equation}
\begin{aligned}
    z'_g &= E_g(z_v),\\
    z''_g = Self\text{-}Attenti&on(W_Q^g z'_g, W_K^g z'_g, W_V^g z'_g),\\
    z_g& = D_g(z''_g).
    \end{aligned}
\end{equation}
where $E_g$ is the compression encoder, and $D_g$ is the decompression function, ensuring $z_g$ matches the original video latent size (detailed designs are in Sec.~\ref{sec:global}).

\hutnew{\noindent\textbf{Time-aware Global-Local Composition.} 
The local and global attention mechanisms yield two latent representations: the local latent $z_l$, capturing fine-grained details, and the global latent $z_g$ , providing semantically coherent global context. To produce videos that are both globally consistent and locally detailed, we introduce a \textbf{global-local fusion module} that combines these representations using a learnable fusion factor $\alpha$.}

\hutnew{During the diffusion process, different denoising timesteps $t$ focus on various video aspects: early timesteps emphasize global structure, while later ones refine details. Thus, the fusion factor $\alpha$ should dynamically adjust with the timestep, shifting focus from global to local information.
To achieve this, we predict $\alpha$ based on timestep $t$. We embed $t$ into a 256-dimensional time feature vector using Sinusoidal Encoding, then project it into a $D$-dimensional fusion factor via an MLP to fuse $z_l$ and $z_g$:}
% This factor is broadcast across $T$ frames and spatial dimensions $W$ and $H$ to form the full fusion map $\alpha \in \mathbb{R}^{B \times T \times H \times W \times D}$:
\begin{equation}
    \begin{aligned}
    \alpha(t) &= MLP(SinEncode(t)), \quad \mathcal{R}^{1}\rightarrow \mathcal{R}^{D}\\
     &z_{fused} = \alpha(t) \cdot z_g + (1 - \alpha(t)) \cdot z_l.\\
    % \alpha(t)&=Repeat(\alpha'(t),T,W,H). \quad \mathcal{R}^{D}\rightarrow \mathcal{R}^{T\times H\times W\times D}
    \end{aligned}
    \label{eq:alpha merging}
\end{equation}
% Finally, we perform fusing $z_l$ and $z_g$ using this adaptive $\alpha$:
% \begin{equation}
% \begin{aligned}
%     z_{fused} &= \alpha(t) \cdot z_g + (1 - \alpha(t)) \cdot z_l,\\
%               % &= \alpha(t) \cdot GA(z) + (1 - \alpha(t)) \cdot LA(z),
%     \end{aligned}
% \end{equation}
% where $LA(\cdot)$ and $GA(\cdot)$ denote the local and global attention mechanisms, respectively. This fusion strategy ensures the model effectively combines local precision with global awareness at each timestep.

% 
\subsection{Spatially-Compressed Global Attention} 
\label{sec:global}

In this section, we \yr{detailedly} introduce our \textit{spatially-compressed global attention} module, \yr{which is} designed to capture global video context while maintaining computational efficiency. The key idea is to compress the spatial dimensions of video latents before performing attention, \yr{so that the self-attention is conducted at a smaller spatial size,} and then decompress them back to the original resolution using spatiotemporal convolution. This reduces the attention cost without sacrificing global modeling capability.

\noindent\textbf{Spatial Compression.} A video can be considered as a sequence of consecutive images, and it is well-known that images can be spatially downsampled to lower resolutions while preserving global semantics at the cost of some local details. Leveraging this property, we propose to spatially compress the video latent by downsampling its width and height by a factor of $k$. This aligns the computational cost of global attention with that of our local attention module.

\hutnew{Specifically, given a video latent $z \in \mathbb{R}^{B \times T \times H \times W \times D}$, we apply a $k\times k$ 2D convolution with stride $k$ along the spatial dimensions $(H, W)$ to obtain a compressed latent $z_c \in \mathbb{R}^{B \times T \times H' \times W' \times D}$, where $H' = \frac{H}{k}$ and $W' = \frac{W}{k}$. }
% The operation is formally defined as:
% \begin{equation}
%     z_c = \text{Conv2D}_{k\timesk,\,\text{stride}=k} (z),
% \end{equation}
% where the convolution is applied independently on each temporal slice of $z$.
To reduce the number of parameters and computational cost in the compression layer, we adopt a channel-wise (i.e., depthwise) convolution mechanism, where each hidden dimension is processed by a separate convolution kernel with a single input and output channel.
Moreover, to ensure training stability at the early stage, we initialize the convolutional kernel weights to be \yr{$1/(k\times k)$}, which initially behaves as average pooling.

\noindent\textbf{Global Attention with Domain-aware LoRA.}  
Once we obtain the compressed video latent $z_c \in \mathbb{R}^{B \times T \times H' \times W' \times D}$, we proceed to apply global self-attention over it. \hut{However, employing both local and global attention mechanisms requires maintaining two attention weights for each, which significantly increases computational overhead.}
% However, directly reusing the original attention weights—designed for local attention—on global latent features can lead to a mismatch in representational semantics. This is because the attention weights $W_Q$, $W_K$, $W_V$, and the feed-forward network (FFN) are inherently biased toward modeling local dependencies.
\hutnew{To address this, we propose a \textbf{domain-aware LoRA} mechanism, which adapts the local attention parameters for global modeling. Specifically, for each projection weight $W \in \{W_Q, W_K, W_V\}$ and the FFN parameters $W_{\text{FFN}}$, we introduce a lightweight, trainable low-rank residual~\cite{hu2022lora} that specializes in global attention. The adapted weight is defined as:}
\begin{equation}
    W^{\text{global}} = W + \Delta W_{\text{LoRA}} = W + A_W B_W,
\end{equation}
where $A_W \in \mathbb{R}^{d \times r}$ and $B_W \in \mathbb{R}^{r \times d}$ are low-rank matrices with rank $r \ll d$, and $d$ is the input/output dimension. The same formulation is applied to $W_{\text{FFN}}$.

\noindent\textbf{Spatiotemporal Decompression.}  
After obtaining the globally modeled compressed latent $z^{\text{global}}_c \in \mathbb{R}^{B \times T \times H' \times W' \times D}$, we need to decompress it back to the original video resolution $T \times H \times W$. 
% Instead of using traditional transposed convolutions—which are known to introduce checkerboard artifacts—we adopt a more stable and artifact-free method based on upsampling followed by \yr{3D} convolution.

\hutnew{Specifically, we first apply bilinear interpolation to upsample the spatial resolution from $H' \times W'$ to $H \times W$. Then, to mitigate the over-smoothing effect caused by interpolation, we apply a convolutional refinement module. Since video frames exhibit not only spatial but also temporal continuity, spatial-only operations may lead to temporal discontinuities. Therefore, we utilize a 3D convolution to perform joint spatio-temporal processing to ensure temporally consistent decompression. The overall process is formulated as:}
\begin{equation}
    z_g = \text{Conv3D}(\text{BilinearUpsample}(z^{\text{global}}_c)),
\end{equation}
\hutnew{where $z_g$ denotes the decompressed global latent, and $\text{Conv3D}$ denotes a 3D convolution operation over the temporal and spatial dimensions. This enables effective restoration of spatial details while preserving temporal coherence.}

\subsection{Cross-window Hierarchical Local Attention}
\label{sec:local}

\hut{\yr{In order to} avoid the quadratic increase in computational complexity as video resolution grows, \yr{we design} \textit{local attention mechanism} \yr{to} partition the video latents into non-overlapping spatial windows \yr{and then conduct self-attention in local windows}. However, %while global attention captures overall correlations across the entire frame, it struggles to 
\yr{this partition makes it difficult to}
model fine-grained relationships at the boundaries between adjacent local windows. To address this issue, we propose \textit{Cross-window Hierarchical Local Attention}, which can effectively model local dependencies within each window and captures interactions between neighboring windows. }

\noindent\textbf{Local Attention.} 
\hutnew{Concretely, we first reshape the video latent $z_v$ into a new tensor of shape ${B \times T \times H \times W \times D}$. We then partition the spatial dimensions $(H, W)$ into $K \times K$ non-overlapping local windows, resulting in a set of local video latent groups $\{v_{i,j}\}_{i=1,j=1}^{K,K}$, where each $v_{i,j} \in \mathbb{R}^{B \times T \times \frac{H}{K} \times \frac{W}{K} \times D}$ corresponds to a spatiotemporal sub-volume of the original video latent:}
\begin{equation}
    v_{i,j} = z_v[:, :, i\cdot \frac{H}{K} : (i+1)\cdot \frac{H}{K}, \, j\cdot \frac{W}{K} : (j+1)\cdot \frac{W}{K}, :] . 
\end{equation}
\hutnew{For each local video latent $v_{i,j}$, we apply self-attention within the it to model the spatiotemporal dependencies:}
\begin{equation}
    v'_{i,j} = Self\text{-}Attention(W_Q^s v_{i,j}, W_K^s v_{i,j}, W_V^s v_{i,j}). 
\end{equation}
\hutnew{By applying self-attention only within each local window, the computational complexity is reduced from $\mathcal{O}((TWH)^2 \cdot D)$ to $\mathcal{O}(K^2 \cdot (\frac{TWH}{K^2})^2 \cdot D) = \mathcal{O}((TWH)^2 \cdot D / K^2)$. As the number of windows increases (i.e., window size decreases), the complexity decreases accordingly. In the extreme case, it reduces the complexity to $\mathcal{O}(TWH \cdot D)$ when each token forms an independent local group, enabling high-resolution video generation at significantly reduced cost.}

After computing self-attention within each local window, we aggregate all locally updated features $\{v'_{i,j}\}$ and restore them to the original video latent %layout. 
\yr{resolution:}
% This is achieved by rearranging the spatial windows back to their original locations in the full-resolution latent tensor. Specifically, the updated local tensors are concatenated along the spatial dimensions to reconstruct the full video latent representation $z_v' \in \mathbb{R}^{B \times T \times H \times W \times D}$:
\begin{equation}
    z_l = Rearrange(\{v'_{i,j}\}_{i,j=1}^{K}). 
\end{equation}
This rearranged $z_l$ preserves the original spatial-temporal resolution of the video while significantly reducing the computation required during attention modeling, which ensures that local details are efficiently captured.

\begin{figure}[t]
    \centering
    \includegraphics[width=0.5\linewidth]{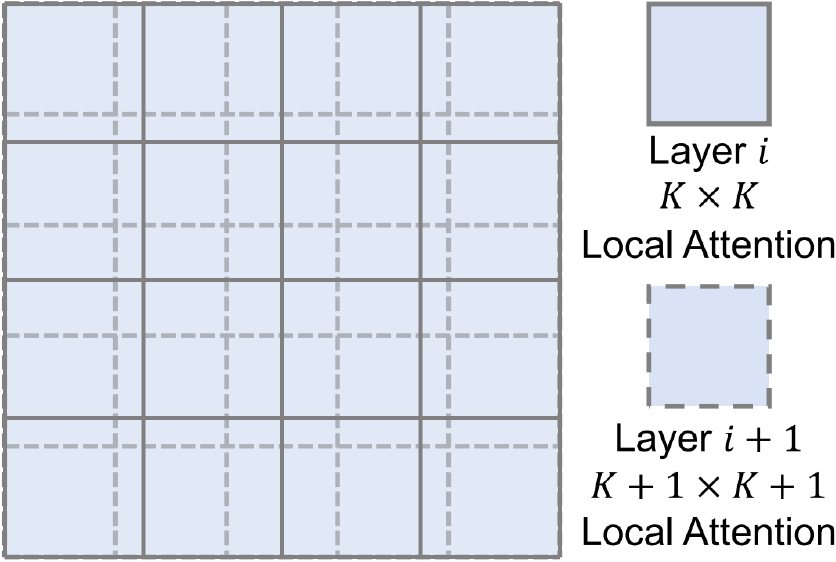}
    \vspace{-0.05in}
    \caption{Cross-window Attention.}
    \label{fig:v1v2}
    \vspace{-0.2in}
\end{figure}

\noindent\textbf{Cross-window Attention.}  Despite incorporating global information modeling, direct communication between local attention windows remains limited, especially at the boundaries, where discontinuities frequently occur. To address this, we propose a \textit{Cross-window Local Attention} to enhance inter-window interaction across local attention \yr{windows}.

Given that the model is composed of multiple layers of transformer blocks, we apply alternating local attention schemes at adjacent layers, \hut{where adjacent layers have \yr{different partition blocks and} window boundaries}. For an even-numbered layer $i$ ($i \bmod 2 = 0$), the spatial domain of the video latent is partitioned into non-overlapping $K \times K$ windows.
For an odd-numbered layer $i$ ($i \bmod 2 = 1$), we apply a shifted window strategy with $(K+1) \times (K+1)$ partitions that partially overlap with the even-layer windows. This cross-window local attention strategy enables hierarchical interaction across neighboring windows between adjacent transformer layers.

\hutnew{As a result, boundary information in the $K \times K$ windows at layer $i$ is propagated through overlapping regions in the $(K+1) \times (K+1)$ windows at layer $i+1$, and vice versa. This enhances continuity across local attention boundaries and improves consistency in the generated outputs. Formally, the attention computation in layer $i$ can be described as:}
\begin{equation}
    z^{(i)}_{cro} = \text{LocalAttn}_{(K + (i \bmod 2)) \times (K + (i \bmod 2))}(z^{(i)}).
\end{equation}
% where $\text{LocalAttn}_{\yr{k} \times \yr{k}}(\cdot)$ denotes window-based self-attention within a $\yr{k} \times \yr{k}$ \yr{partitioned} region. This \yr{cross-window} cross-layer design allows edge information in local windows to be effectively exchanged and fused over layers.

\noindent\textbf{Hierarchical Local Attention.}  
While the proposed cross-window local attention enhances information exchange across adjacent local attention windows, it may still \yr{be} hard to capture fast-moving small objects, which can simultaneously span multiple local windows between frames. In such cases, the limited overlapping in cross-window attention is insufficient, and global attention lacks the resolution to model fine-grained local details. 
\hut{To address this, we introduce a \textit{Hierarchical Local Attention} (HLA) mechanism, which divides the full attention into $(K/2) \times (K/2)$ \yr{coarse} windows \yr{(each twice the size as the local window)}, and performs local attention within each \yr{coarse} window at an intermediate scale. This approach effectively compensates for the inability of global attention to capture fine-grained details, while also overcoming the limited receptive field inherent in conventional local attention mechanisms. }

Specifically, we first compress the latent features within each local window using a strategy similar to our spatial-compressed global attention. The local latent $z_c^{hla}$ within each %$(K/2) \times (K/2)$ region 
\yr{coarse window of size $\frac{2H}{K} \times \frac{2W}{K}$}
is downsampled via strided convolution.
% \begin{equation}
%     z_c^{hla} = \text{Conv}_{k=3, s=2}(z_v).
% \end{equation}
To effectively model the hierarchical attention, we apply a \textit{domain-aware LoRA} adaptation to the pretrained attention weights (including $W_Q$, $W_K$, $W_V$, and FFN) to ensure they are appropriately adapted for hierarchical attention computation:
\begin{equation}
    W_{\text{hla}} = W_{\text{local}} + \Delta W^{\text{HLA}},
\end{equation}
where $\Delta W^{\text{HLA}}$ is the domain-specific LoRA adaptation for hierarchical attention.

\begin{figure*}[t]
    \centering
     \includegraphics[width=0.9\textwidth]{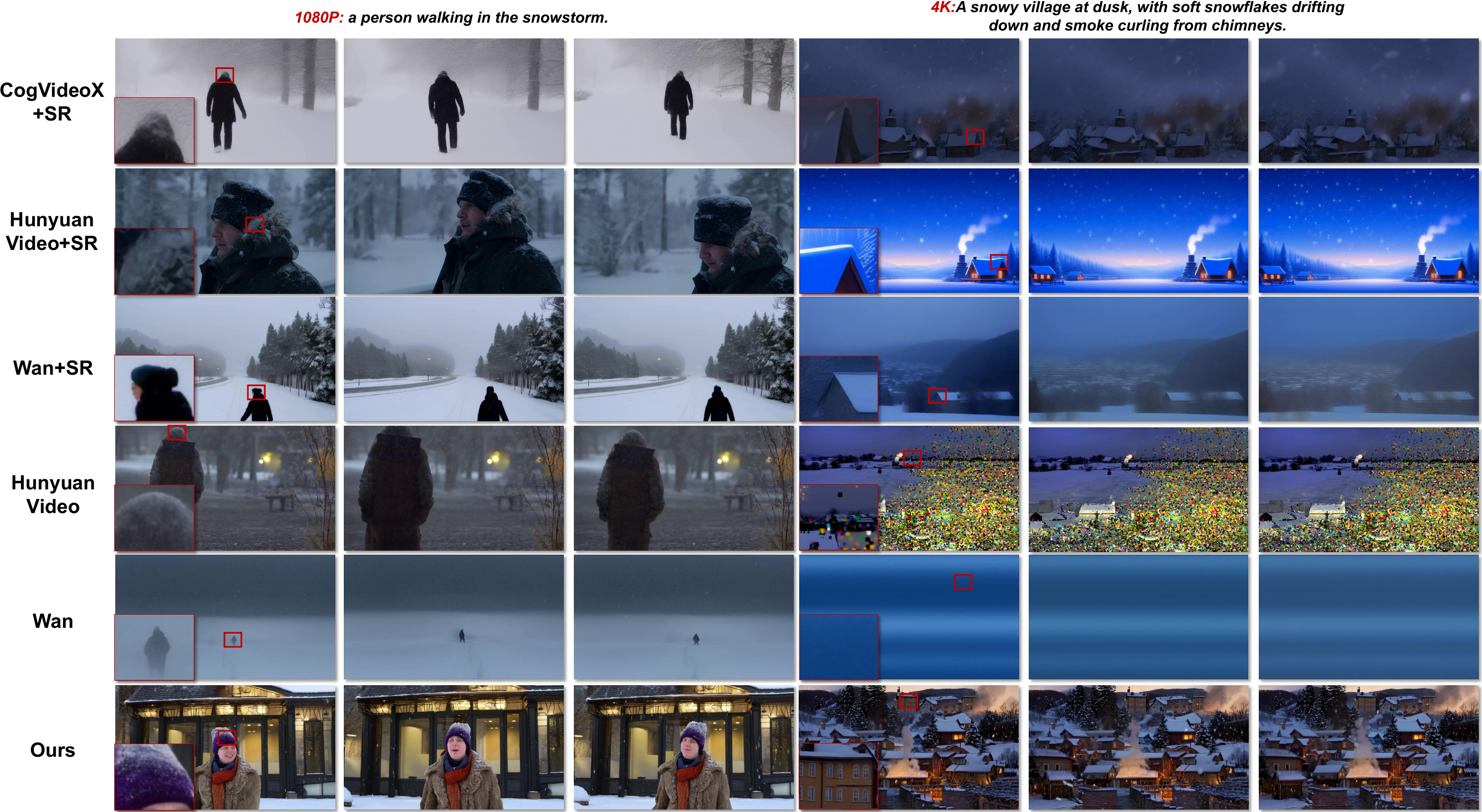}
    \caption{ Comparison results of existing state-of-the-art video generation methods on 1080P video generation. The \textcolor{red}{red} boxes highlight zoomed-in regions, where our model produces the clearest high-resolution videos with the most fine-grained details.}
    \label{fig:compare on 1k and 4k}
\end{figure*}

Similar to the cross-window local attention design, we employ an alternating shift mechanism between adjacent transformer layers to ensure information flow across hierarchical attention windows. That is, for layer $i$, hierarchical attention is computed with non-overlapping $(K/2) \times (K/2)$ windows; for layer $i+1$, \yr{we partition the spatial domain into} $(K/2 + 1) \times (K/2 + 1)$ \yr{non-overlapping} windows, %enabling overlap 
\yr{making the windows of adjacent layers intersect with each other} 
and thus \yr{enabling} boundary information propagation. 
The attention operation at each layer can be described as:
\begin{equation}
    z^{(i)}_{hla} = HierAttn_{(\frac{K}{2} + (i \bmod 2)) \times (\frac{K}{2} + (i \bmod 2))}(z^{(i)}),
\end{equation}
where $HierAttn_{\yr{k} \times \yr{k}}(\cdot)$ denotes attention over a $\yr{k} \times \yr{k}$ \yr{partitioned} hierarchical window.

\hutnew{This hierarchical structure, combined with cross-layer shift design and domain-aware adaptation, enables efficient fine-grained motion modeling of fast-moving small objects and enhances the robustness of local attention modeling in dynamic video scenes.  To fuse the results from both the \textit{Cross-window Local Attention} $z_{cro}$ and the \textit{Hierarchical Local Attention} $z_{hla}$, we employ a time-aware alpha $\alpha_{local}$ to fuse the two results, which is the same as the \textit{Time-aware Global-Local Composition}.}

\section{Experiments}
\label{sec:experiment}

\subsection{Implementation Details}

\noindent\textbf{Baselines.} We compare our model with state-of-the-art methods, including Wan~\cite{wang2025wan}, HunyuanVideo~\cite{hunyuanvideo}, and CogVideo-X~\cite{yang2024cogvideox}. For each method, we generate two sets of videos: 1) one by directly generating videos at the target resolution, and 2) the other by first generating videos at the default resolution and then applying a super-resolution method~\cite{zhang2024realviformer} to upscale them to the target size. Note that CogVideoX cannot support HD video generation; therefore, we directly combine it with video super-resolution.

\noindent\textbf{Evaluation Metrics.} Conventional metrics such as FVD~\cite{unterthiner2018fvd} are inadequate for evaluating the quality of high-resolution video generation, as they rely on pretrained low-resolution video encoders that fail to capture high-resolution features. To address this limitation, we introduce three novel metrics specifically designed for high-resolution video evaluation: \textbf{\textit{1)} HD-FVD} measures the similarity between generated and real high-resolution videos, while \textbf{\textit{2)} HD-MSE} and \textbf{\textit{3)} HD-LPIPS} assess the fine-grained pixel-level and semantic-level~\cite{zhang2018lpips} details of the generated videos, respectively. 
% Lower values indicate better performance, and 
 Additional CLIP score~\cite{clip} and temporal consistency~\cite{huang2024vbench} are included for a more comprehensive evaluation. \hutnew{Further details and more Vbench metrics ~\cite{huang2024vbench} are provided in the appendix.}

\subsection{Comparison Results}
\begin{figure}[t]
    \centering
    \includegraphics[width=0.47\textwidth]{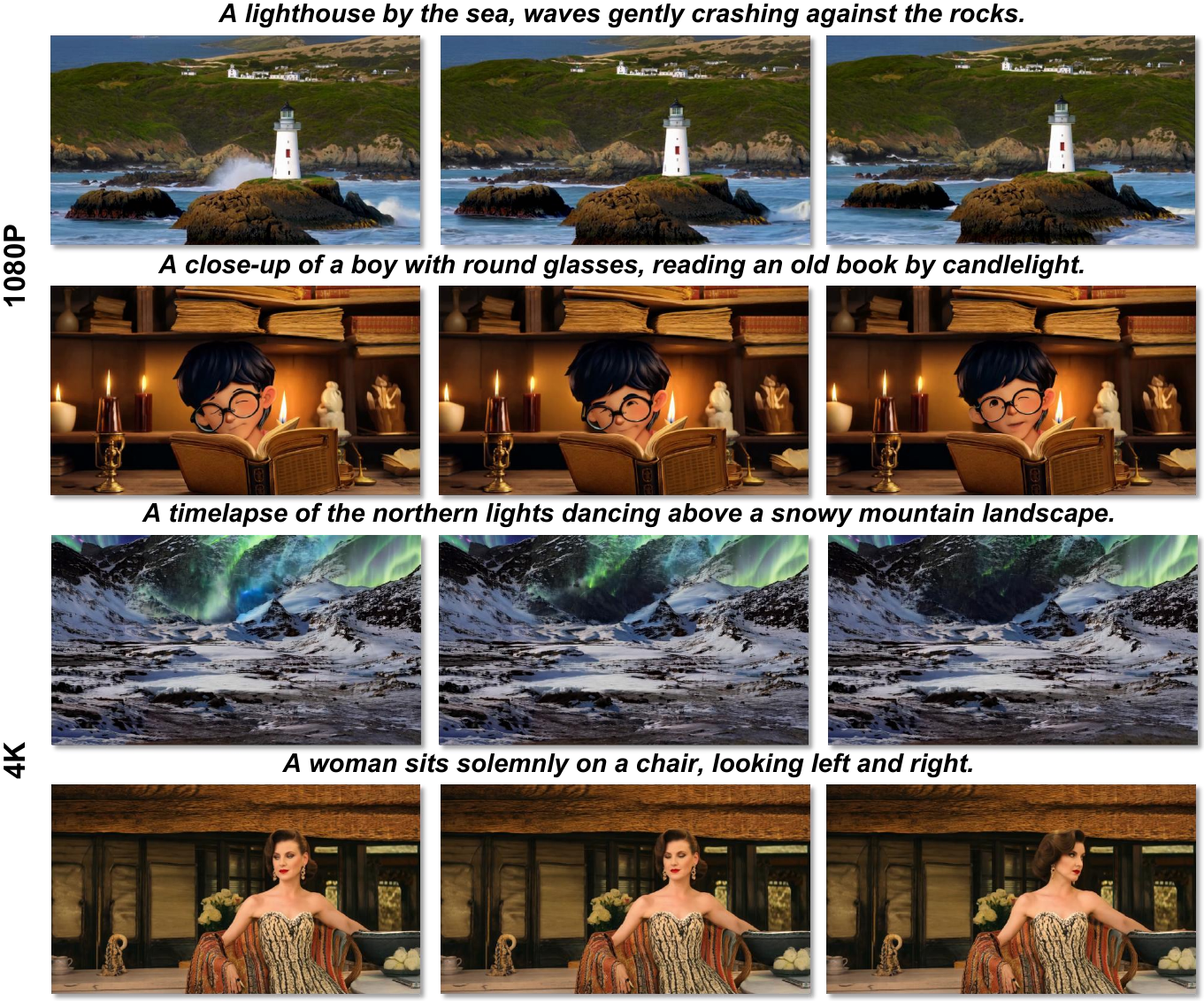}
    \vspace{-0.1in}
    \caption{More generated HD videos (1080P \& 4K).}
    \vspace{-0.1in}
    \label{fig:more results}
\end{figure}

\textbf{Qualitative Comparison.} We compare our model with state-of-the-art methods on both 1080P and 4K video generation tasks. The comparison results are shown in Fig.~\ref{fig:compare on 1k and 4k}. As can be seen, the Wan model is unable to directly generate 1080P videos, resulting in blurry outputs with little to no semantic content. HunyuanVideo is capable of generating 1080P videos, but often produces results with incorrect semantics that are inconsistent with the given prompt. Methods that combine super-resolution models can generate text-aligned videos; however, the outputs after super-resolution tend to be overly smooth and lack fine details. Among these, only HunyuanVideo+SR produces relatively good results, but the level of detail is still significantly lower than that of our model, as highlighted in the zoomed-in red boxes. Therefore, our model is able to generate high-resolution videos with fine-grained details while faithfully following the given prompt, demonstrating its superior performance in high-resolution video generation. \hutnew{Moreover, additional results generated by our model are presented in Fig.~\ref{fig:more results}. \yrn{It can be seen that} our model consistently produces high-quality HD videos across various prompts.}

\begin{table}[t]
\renewcommand{\arraystretch}{1.0}
\setlength\tabcolsep{6.0pt}
\resizebox{0.47\textwidth}{!}{
\begin{tabular}{c|c|c|ccccc}
\toprule
\makecell[c]{Reso-\\lution}             & Method       & SR& \makecell[c]{HD-\\FVD $\downarrow$} & \makecell[c]{HD-\\MSE} $\uparrow$ & \makecell[c]{HD-\\LPIPS $\uparrow$} & CLIP-L $\uparrow$ & \makecell[c]{Temporal\\ Consis $\uparrow$}  \\
\midrule
\multirow{6}{*}{1080P} & CogVideoX     & \ding{52} &   394.82     & 97.21  & 0.3060     & 0.2834 & 0.9468                \\
                       & HunyuanV & \ding{52} &    238.75    & 126.68 & 0.3590     & 0.2883 & 0.9614               \\
                       & Wan          & \ding{52} &   309.10     & 163.86 & 0.3499    & 0.2747 & 0.9750              \\ \cline{2-8}
                       & HunyuanV & \ding{56} &     237.89    & 207.68 & 0.4911     & 0.2636 & 0.9752           \\
                       & Wan          & \ding{56} &   821.54        & 42.93  & 0.4290     & 0.2528 & 0.9768           \\
                       & \textbf{Ours}         & \ding{56} &     \textbf{214.12}    & \textbf{390.19} & \textbf{0.5455}     & 0.2654$^*$ & \textbf{0.9827}          \\
                       \midrule
\multirow{6}{*}{4K}    & CogVideoX     & \ding{52} &   574.10     &      68.94  &	0.2645          & 0.2436 & 0.9449             \\

                       & HunyuanV & \ding{52} &     453.41   & 276.76 & 0.4066     & 0.2576 & 0.9684                 \\
                       & Wan          & \ding{52} &    471.56   & 77.67  & 0.2782     & 0.2455 & 0.9697               \\ \cline{2-8}
                       & HunyuanV & \ding{56} &    805.42    & 102.36	      &   0.3858        &    0.2151    &    0.9679             \\
                       & Wan          & \ding{56} &    1272.08    &    29.45	    &   0.4270       &  0.2123     &   0.9705               \\
                       & \textbf{Ours}         & \ding{56} &     \textbf{424.61}   &     \textbf{386.01}   &  \textbf{0.6450}        &  0.2444$^*$      &  \textbf{0.9710}                 \\ 
                       \bottomrule          
\end{tabular}}
\caption{Quantitative comparisons. \zjn{Our UltraGen demonstrates superior high-quality HD video generation capabilities}. \textbf{Bold} indicates the best performance and $^*$ indicates the best performance among all the non-SR methods.}
\label{tab:quantitative comparison}
\vspace{-0.1in}
\end{table}

\textbf{Quantitative Comparison.} We compare our method with state-of-the-art approaches in Tab.~\ref{tab:quantitative comparison}. For HD evaluation metrics, our model achieves the lowest HD-FVD scores on both 1080P and 4K video generation, indicating superior quality and diversity in the generated videos. Furthermore, we obtain the best HD-MSE and HD-LPIPS, demonstrating that our generated videos contain the most fine-grained details and validating the effectiveness of our HD video generation ability. Our model also achieves the best temporal consistency, which demonstrates the smoothness of the generated videos and the coherence across frames.
In terms of prompt following, we observe that directly generating HD videos without super-resolution leads to a relatively lower CLIP score due to the difficulty in high-resolution video generation. Since our model is based on Wan 1.3B, its CLIP score cannot surpass that of Wan+SR. Nevertheless, we still achieve the best CLIP score among methods that natively generate high-resolution videos, highlighting the strong prompt-following capability of our model.

\textbf{Time Comparison.} Finally, we compare the inference time of our model with HunyuanVideo and Wan at different resolutions, as shown in Tab.~\ref{tab:time comparison}. Our model achieves a 2.7$\times$ speedup for 1080P generation and a \textbf{4.78$\mathbf{\times}$ speedup} for 4K generation compared to the baseline Wan model, demonstrating the high efficiency of our approach for high-resolution video generation.

% \begin{table*}[t]
% \resizebox{0.9\linewidth}{!}{
% \begin{tabular}{ccccccccccccc}
% \toprule
% Resolution      & \multicolumn{6}{c}{1080P}                                          & \multicolumn{6}{c}{4K}                                             \\
% Method          & CogVideo+SR & HunyuanVideo & HunyuanVideo+SR & Wan & Wan+SR & Ours & CogVideo+SR & HunyuanVideo & HunyuanVideo+SR & Wan & Wan+SR & Ours \\ \midrule
% HD-FVD          &             &              &                 &     &        &      &             &              &                 &     &        &      \\
% HD-MSE          &             &              &                 &     &        &      &             &              &                 &     &        &      \\
% HD-LPIPS        &             &              &                 &     &        &      &             &              &                 &     &        &      \\
% Temporal Consis &             &              &                 &     &        &      &             &              &                 &     &        &      \\
% Dynamic Degree  &             &              &                 &     &        &      &             &              &                 &     &        &      \\
% \bottomrule
% \end{tabular}}
% \end{table*}

\begin{table}[t]
\renewcommand{\arraystretch}{1.0}
\setlength\tabcolsep{6.0pt}
\resizebox{0.48\textwidth}{!}{
\begin{tabular}{c|ccc|c}
\toprule
Resolution & HunyuanVideo & Wan      & UltraGen (Ours) & Speedup (Ours) \\ \midrule
1080P      & 43 min       & 35 min   & \textbf{13 min}          & $\times$2.69            \\
4K         & 11h 36min    & 8h 46min & \textbf{1h 50min}        & $\times$4.78      \\    \bottomrule
\end{tabular}}
\vspace{-0.1in}
\caption{Comparison of inference time. Our model archives a \textbf{4.78 $\mathcal{\times}$ speedup} compared to the baseline Wan model.}
\label{tab:time comparison}
\vspace{-0.1in}
\end{table}

\subsection{Ablation Studies}

\begin{figure}[t]
    \centering
    \includegraphics[width=0.47\textwidth]{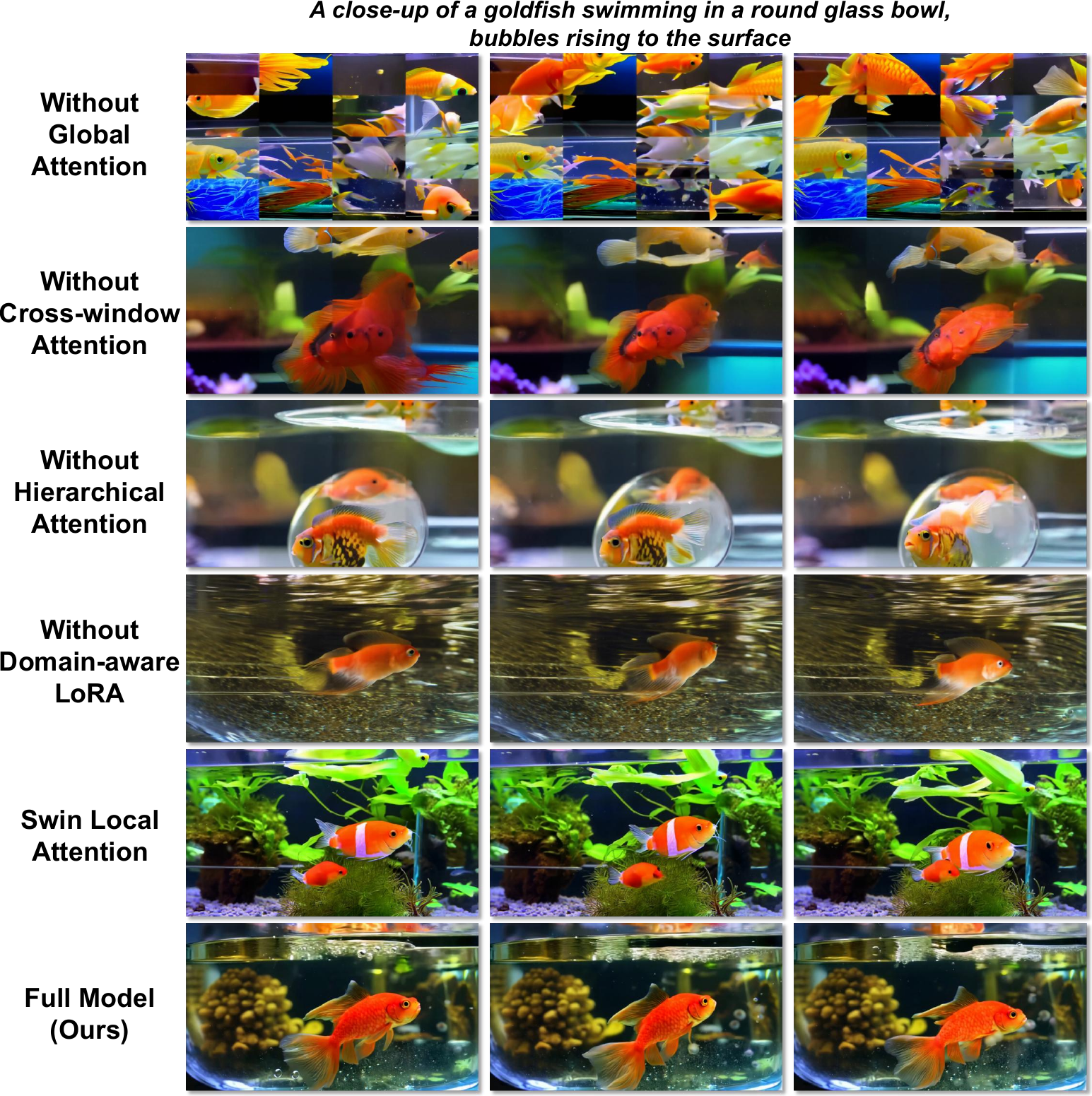}
    \vspace{-0.1in}
    \caption{ \hutnew{Ablation study on the proposed modules.}}
    \vspace{-0.1in}
    \label{fig:ablation study}
\end{figure}

We conduct ablation studies on five variants: (1) without global attention, (2) without hierarchical attention, (3) without domain-aware LoRA, (4) without cross-window local attention, and (5) employing Swin-Attention~\cite{liu2021swin} for local attention modeling. As shown in Fig.~\ref{fig:ablation study}, the model without global attention tends to generate disjoint content, exemplified by the isolated 16 golden fishes in the rightmost case. Models lacking either cross-window local attention or hierarchical attention can capture global relationships only coarsely and still exhibit inconsistencies at window boundaries. The model without domain-aware LoRA alleviates boundary inconsistency but suffers from reduced generation quality, producing somewhat blurry results. This is due to the limited capacity of a single set of attention weights to model three distinct attention mechanisms (global, local, and hierarchical). Moreover, when replacing hierarchical cross attention with Swin-Attention for local attention modeling, we observe that although adjacent windows can be connected smoothly, Swin-Attention struggles to effectively capture hierarchical features. As a result, the model often generates semantically inconsistent content across windows. For example, it may produce two goldfish in adjacent windows where only one should appear, indicating a lack of semantic coherence. In contrast, the full model generates high-quality videos, effectively resolves boundary inconsistencies, and captures global semantics well, validating the effectiveness of all our proposed modules. \hutnew{More quantitative ablation studies are shown in the appendix.}

% \subsection{More Visualization Results}

% To provide a more comprehensive evaluation of our model, we present additional generated videos in Fig.~\ref{fig:more visulizations}. \yr{It can be seen that} our model consistently produces high-quality HD videos with fine-grained details at both 1080P and 4K resolutions. Furthermore, the generated videos closely follow the input prompts while maintaining strong temporal consistency and overall video quality.
\section{Conclusion}

In this work, we propose UltraGen, a novel framework for efficient, end-to-end native high-resolution video generation. By leveraging a hierarchical dual-branch attention architecture, UltraGen effectively decouples local and global attention, enabling the synthesis of high-fidelity regional details while maintaining overall semantic consistency. Our spatially compressed global modeling and hierarchical cross-window local attention mechanisms further reduce computational complexity, making high-resolution video generation (up to 4K) feasible for both training and inference. Extensive experiments demonstrate that UltraGen not only scales pre-trained low-resolution models to 1080P and 4K resolutions, but also consistently outperforms existing state-of-the-art methods and super-resolution pipelines in both qualitative and quantitative evaluations. 
% We believe UltraGen paves the way for practical and scalable high-resolution video generation, opening new possibilities for applications in content creation, digital entertainment, and beyond.

{
    \small
    \bibliographystyle{ieeenat_fullname}
    \bibliography{main}
}

\clearpage
\appendix

\section{Overview}
This supplementary material consists of:

\begin{itemize}
    \item Efficiency analysis (Sec.~\ref{sec:efficiency analysis});
    \item More implementation details (Sec.~\ref{sec:more implementation details});
    \item High-resolution video evaluation metrics (Sec.~\ref {sec:evaluation metrics});
    \item Evaluation on Vbench (Sec.~\ref{sec:vbench_evaluation});
     \item Quantitative ablation studies (Sec.~\ref{sec:quantitative ablation study});
     \item More visualization results (Sec.~\ref{sec: more visualization});
    \item Limitations (Sec.~\ref{sec:limitations}).
\end{itemize}

% For more experiment results, please refer to the demo video.

\section{Efficiency Analysis}
\label{sec:efficiency analysis}

Since the primary computational cost of full attention lies in the calculation of the attention map, we approximate the overall computational complexity by analyzing the complexity of the attention map itself. The original full attention mechanism has a computational complexity of $\mathcal{O}((TWH)^2 D)$.

We consider the case without cross-window connections for ease of analysis (note that introducing cross-window connections increases the effective $K$, which can further reduce the computational cost to some extent), where the input is partitioned into $K \times K$ local windows. The computational complexity of our local attention is then $\mathcal{O}(K^2 \cdot (\frac{TWH}{K^2})^2 \cdot D) = \mathcal{O}((TWH)^2 D / K^2)$.

In addition to local attention, our model also incorporates global attention and hierarchical attention. However, by leveraging a global latent compression mechanism, we ensure that the attention map sizes for these two modules are consistent with that of local attention. Specifically, the computational complexity for global attention is $\mathcal{O}((\frac{TWH}{K^2})^2 D)$, and for hierarchical attention, it is $\mathcal{O}((\frac{K}{2})^2 \cdot (\frac{TWH}{K^2})^2 D) = \mathcal{O}((TWH)^2 D / (4 K^2))$.

Therefore, the total computational complexity \yr{is} expressed as:
\begin{equation}
\begin{aligned}
    \yr{T(n)}&= \mathcal{O}\left(\frac{(TWH)^2 D}{K^2}\right) + 
    \mathcal{O}\left(\frac{(TWH)^2 D}{K^4}\right) \\
    &\quad + \mathcal{O}\left(\frac{(TWH)^2 D}{4K^2}\right) \\
    &=\mathcal{O}\left(\frac{5(TWH)^2 D}{4K^2} + \frac{(TWH)^2 D}{K^4}\right),
\end{aligned}
\end{equation}

Then, the speedup ratio compared to the standard complexity $\mathcal{O}((TWH)^2 D)$ is:
\begin{equation}
\begin{aligned}
\text{Speedup} 
&= \frac{(TWH)^2 D}{\frac{5}{4K^2}(TWH)^2 D + \frac{1}{K^4}(TWH)^2 D}\\
&= \frac{1}{\frac{5}{4K^2} + \frac{1}{K^4}}\\
&= \frac{4K^4}{5K^2 + 4}
\end{aligned}
\end{equation}
where $K=4$ is used in our experiments, resulting in an approximate $12$-fold speedup. However, in practice, the actual speedup is somewhat lower than $12$ due to the additional computation required for generating queries, keys, and values in the global and hierarchical attention modules. Nevertheless, as the resolution increases and the attention map computation becomes the dominant cost, the observed speedup approaches the theoretical value of $12$.

\section{More Implementation Details}
\label{sec:more implementation details}

\textbf{Training and Inference Details.} We perform full fine-tuning on the pretrained Wan 1.3B model \cite{wang2025wan}, integrating domain-aware LoRA with a rank of 64 for both global and hierarchical attention mechanisms. The training process utilizes the UltraVideo dataset \cite{xue2025ultravideo}, which comprises 42,000 4K-resolution videos, and is conducted over 50 epochs. Training is executed on 32 H20 GPUs with a batch size of 32 and a learning rate of $1 \times 10^{-4}$. For the 1080P video generation model, we fix the number of frames at 81, following the official configuration of Wan. For the 4K model, due to GPU memory constraints, we are only able to train the model with 29 frames.
For inference, we employ 30 denoising steps and set the classifier-free guidance scale to 5.0.

\begin{table*}[t]
\centering
\renewcommand{\arraystretch}{1.0}
\setlength\tabcolsep{4.5pt}
\resizebox{0.8\textwidth}{!}{
\begin{tabular}{c|c|c|cccccc}
\toprule
Resolution & Method & SR & 
\makecell[c]{Subject \\Consistency $\uparrow$} & 
\makecell[c]{Background \\Consistency $\uparrow$} & 
\makecell[c]{Motion \\Smoothness $\uparrow$} & 
\makecell[c]{Aesthetic \\Quality $\uparrow$} & 
\makecell[c]{Imaging \\Quality $\uparrow$} & 
\makecell[c]{Average $\uparrow$} \\
\midrule
\multirow{4}{*}{1080P} 
& CogVideoX   & \ding{52} & 0.9456 & 0.9592 & 0.9901 & 0.5138 & 0.5771 & 0.7972 \\
& HunyuanVideo    & \ding{56} & 0.9796 & 0.9839 & 0.9967 & 0.5892 & 0.6237 & \underline{0.8346} \\
& Wan        & \ding{56} & 0.9770 & 0.9762 & 0.9967 & 0.4317 & 0.4529 & 0.7669 \\
& \textbf{Ours} & \ding{56} & {0.9771} & {0.9777} & {0.9961} & 0.5819 & {0.7350} & \textbf{0.8536} \\
\midrule
\multirow{4}{*}{4K}
& CogVideoX   & \ding{52} & 0.9472 & 0.9575 & 0.9895 & 0.5072 & 0.5708 & \underline{0.7944} \\
& HunyuanVideo    & \ding{56} & 0.9964 & 0.9967 & 0.9979 & 0.3973 & 0.4402 & 0.7657 \\
& Wan        & \ding{56} & 0.9466 & 0.9764 & 0.9952 & 0.2877 & 0.3735 & 0.7159 \\
&\textbf{Ours} & \ding{56} & 0.9854 & 0.9894 & 0.9933 & 0.5787 & 0.6832 & \textbf{0.8460}\\
\bottomrule
\end{tabular}
}
\caption{Quantitative comparison on selected methods using VBench metrics. \textbf{Bold} denotes the best score.}
\label{tab:vbench_metrics}
\end{table*}

\section{High-resolution Video Evaluation Metrics.}
\label{sec:evaluation metrics}

\textbf{HD-FVD:} The standard FVD metric~\cite{unterthiner2018fvd} utilizes the I3D network~\cite{carreira2017i3d} to extract video features, which involves resizing input videos to a low resolution ($H_l\times W_l$) prior to feature extraction and comparison. To enable evaluation at high resolutions, we propose HD-FVD, which decomposes high-resolution videos into patches of size $H_l\times W_l$. Features are then extracted from these patches using the pretrained I3D network, thereby preserving high-resolution information. The Fréchet Distance is subsequently computed between the features of generated and reference video patches.

\textbf{HD-MSE:} High-resolution videos inherently contain fine details that are absent in their low-resolution counterparts. To quantitatively assess the preservation of such details, we first downsample the videos by set of factors of $\{2^k\}$, resulting in a set of downsampled videos $\{v_{D,2^k}\}$. Each downsampled video is then upsampled back to the original resolution, and the mean squared error (MSE) is computed with respect to the original video. This process is formalized as:
    \begin{equation}
        \mathrm{HD\text{-}MSE} = \sum_k \|v - v_{D,2^k}\|
        \label{eq:HD-MSE}
    \end{equation}
    A higher HD-MSE indicates that more fine details are lost during downsampling, thereby reflecting the presence of high-quality, high-resolution content in the generated videos. In our experiment, we enumerate $k$ from $3$ to $5$ (corresponds downsample factor $8$, $16$, and $32$) to compute the HD-MSE.

\textbf{HD-LPIPS:} Analogous to HD-MSE, HD-LPIPS evaluates the preservation of fine-grained semantic details in high-resolution videos. Here, the MSE in Eq.~\ref{eq:HD-MSE} is replaced with the LPIPS metric~\cite{zhang2018lpips}, which is more sensitive to perceptual differences:
    \begin{equation}
        \mathrm{HD\text{-}LPIPS} = \sum_k LPIPS(v - v_{D,2^k}),
        \label{eq:HD-LPIPS}
    \end{equation}
    where we use $k=\{3,4,5\}$ to compute HD-LPIPS.

\section{Evaluation on Vbench}
\label{sec:vbench_evaluation}
To further demonstrate the effectiveness of our approach, we conduct comparisons with several state-of-the-art methods using the VBench evaluation framework~\cite{huang2024vbench}. All methods are evaluated under identical resolution and prompt settings. VBench offers a standardized and comprehensive suite of metrics—including subject consistency, background consistency, motion smoothness, aesthetic quality, and imaging quality—enabling a thorough assessment of video generation quality. It is important to note that VBench is not designed for high-resolution generation; thus, videos must be resized to the standard resolution of the pretrained models for evaluation. As a result, super-resolution-based methods such as Wan~\cite{wang2025wan} and HunyuanVideo~\cite{hunyuanvideo} are not included in our comparison. When their high-resolution outputs are downsampled to the standard resolution for VBench evaluation, the assessment essentially reflects the performance of the base models (i.e., HunyuanVideo and Wan) rather than their high-resolution generation capabilities, which would not provide a fair comparison in the high-resolution video generation setting.

Table~\ref{tab:vbench_metrics} reports the quantitative results on the generated videos from different methods. It can be seen that our method achieves the highest overall average score in both 1080P and 4K resolution, demonstrating a balanced and robust performance across diverse aspects of video quality. Notably, our approach attains the best \textit{Imaging Quality} score, reflecting its strong ability to mitigate low-level distortions such as blur and noise in high-resolution frames. While it does not outperform all competitors on every individual metric, it consistently ranks near the top across all categories. In comparison, methods that directly generate high-resolution videos, such as Wan and HunyuanVideo, show significantly lower scores on perceptual quality metrics like \textit{Aesthetic Quality} and \textit{Imaging Quality}, indicating challenges in preserving fine details and reducing artifacts at scale. These results validate the effectiveness of our model in producing high-fidelity, temporally coherent videos with fewer visual distortions.

\begin{table*}[t]
\centering

\renewcommand{\arraystretch}{1.0}
\resizebox{0.98\textwidth}{!}{
\begin{tabular}{c|cc|cccccc}
\toprule
Method & HD-FVD $\downarrow$ & CLIP-L $\uparrow$ & \makecell[c]{Subject \\ Consistency$\uparrow$} & \makecell[c]{Background \\ Consistency$\uparrow$} & \makecell[c]{Motion \\ Smoothness$\uparrow$} & \makecell[c]{Aesthetic \\ Quality$\uparrow$} & \makecell[c]{Imaging \\ Quality$\uparrow$} & Average$\uparrow$\\
\midrule
without global attention &328.98&0.2302& 0.9680 & 0.9692 & 0.9919 & 0.4489 & 0.6386 & 0.8033 \\
without cross-window attention &419.15&0.2488& 0.9720 & 0.9725 & 0.9929 & 0.4369 & 0.6964 & 0.8141 \\
without hierarchical attention &376.49&0.2581& \textbf{0.9800} & \textbf{0.9821} & 0.9943 & 0.5400 & 0.6784 & 0.8350 \\
without domain-aware LoRA & \underline{284.08}&\underline{0.2603} & \underline{0.9790} & \underline{0.9791} & \underline{0.9948} & \underline{0.5541} & \textbf{0.7424} & \underline{0.8499} \\
swin local attention & 458.93&0.2548&0.9789 & 0.9756 & 0.9943 & 0.5308 & 0.7228 & 0.8405 \\
\textbf{full model (ours)} & \textbf{214.12}& \textbf{0.2654}& 0.9771 & 0.9777 & \textbf{0.9961} & \textbf{0.5819} & \underline{0.7350} &\textbf{0.8536} \\
\bottomrule
\end{tabular}}
\caption{Quantitative ablation study.}
\label{tab:ablation_study_HD}
\end{table*}

\begin{figure*}[t]
    \centering
    \includegraphics[width=1.0\textwidth]{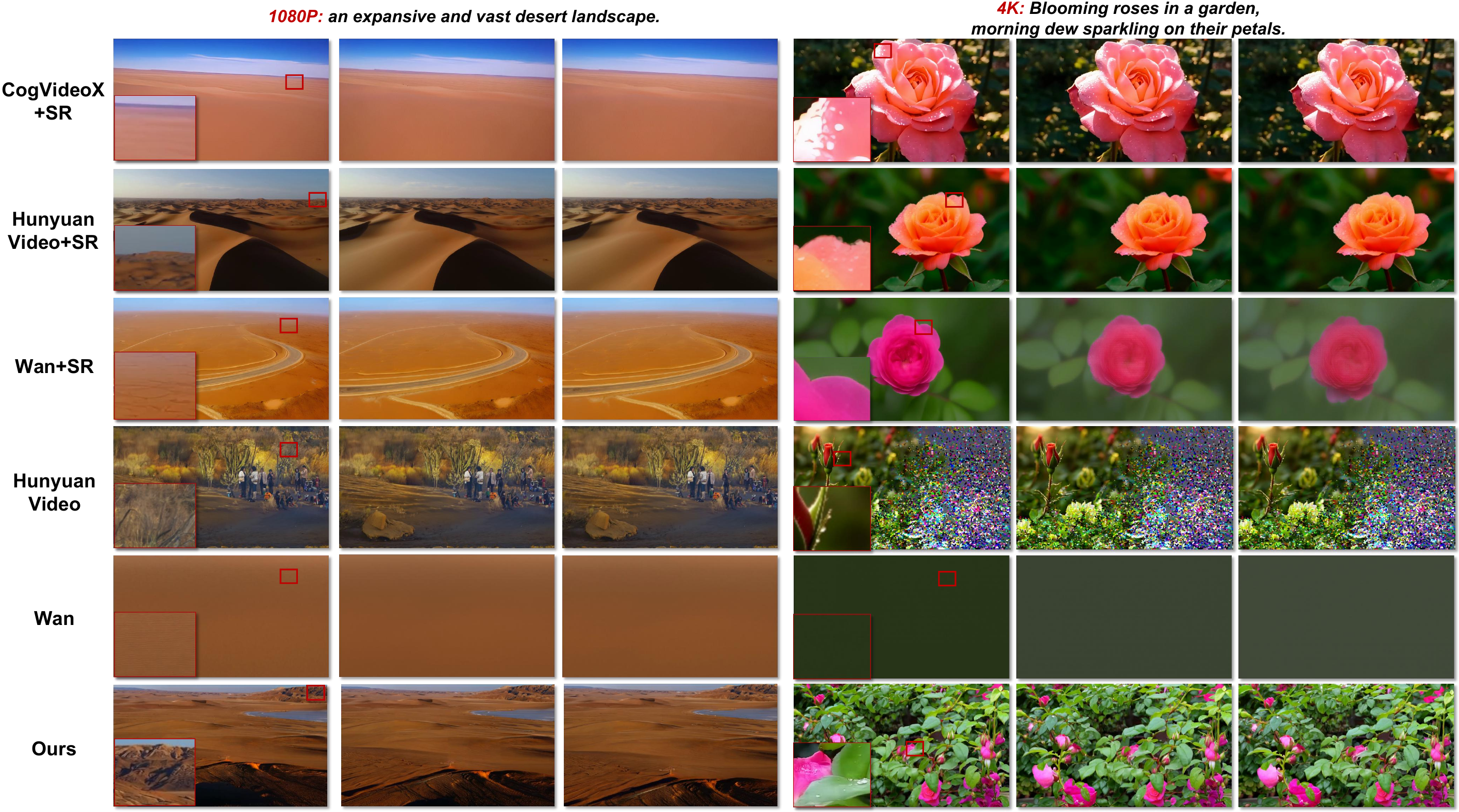}
    \caption{More Qualitative comparisons between our UltraGen and the existing HD video generation methods.}
    \label{fig:compare on 1k and 4k suppl}
\end{figure*}

\section{Quantitative Ablation Studies.}
\label{sec:quantitative ablation study}

In the main paper, we have demonstrated the effectiveness of each proposed module in UltraGen. To provide a more comprehensive and rigorous evaluation, we present additional quantitative ablation studies in this section, examining five ablated variants: (1) without global attention, (2) without hierarchical attention, (3) without domain-aware LoRA, (4) without cross-window local attention, and (5) replacing our local attention module with Swin-Attention~\cite{liu2021swin}. Quantitative comparisons are reported in Table~\ref{tab:ablation_study_HD} using the HD-FVD, CLIP-L, and VBench metrics.
Our model achieves the best performance on both HD-FVD and CLIP-L, indicating superior high-definition generation quality. Furthermore, with respect to the VBench metrics, our model attains the highest scores in motion smoothness and aesthetic quality, as well as the second-best imaging quality, resulting in the highest overall average VBench score. Notably, the variants without global attention or cross-window local attention exhibit severe boundary inconsistencies, leading to the lowest aesthetic and imaging quality. Both hierarchical attention and domain-aware LoRA contribute to improved generation quality; omitting either results in a moderate decrease in performance. Compared to the Swin-Attention variant, our hierarchical cross-layer mechanism demonstrates superior performance in high-resolution video generation.
In summary, our model achieves state-of-the-art HD video generation performance, validating the effectiveness of each proposed module.

% \begin{table*}[t]
% \centering
% \renewcommand{\arraystretch}{1.0}
% % \setlength\tabcolsep{6.0pt}
% % \resizebox{0.7\textwidth}{!}{
% \begin{tabular}{c|ccccc}
% \toprule 
% Method       & \makecell[c]{HD-\\FVD $\downarrow$} & \makecell[c]{HD-\\MSE} $\uparrow$ & \makecell[c]{HD-\\LPIPS $\uparrow$} & CLIP-L $\uparrow$ & \makecell[c]{Temporal\\ Consis $\uparrow$}  \\
% \midrule 
% without global attention & 328.98 & 287.28 & 0.4490 & 0.2302 & - \\
% without cross-window attention & 419.15 & 420.09 & 0.5411 & 0.2488 & - \\
% without hierarchical attention & 376.49 & 441.38 & 0.4256 & 0.2581 & - \\
% without domain-aware LoRA & 284.08 & 562.15 & 0.6219 & 0.2603 & - \\
% swin local attention & 458.93 & 530.76 & 0.5590 & 0.2548 & - \\
% \textbf{full model(ours)}         &     214.12    & 390.19 & 0.5455     & 0.2654 & 0.9827          \\
% \bottomrule
% \end{tabular}
% % }
% \caption{Ablation study evaluating the impact of different attention mechanisms and components in our video generation framework.}
% \label{tab:ablation_study_HD}
% \end{table*}

\section{More Visualization Results.}
\label{sec: more visualization}

\textbf{More qualitative comparisons.} In this section, we present additional qualitative comparisons between our UltraGen model and several baseline methods, including CogVideoX~\cite{yang2024cogvideox}+SR, HunyuanVideo~\cite{hunyuanvideo}+SR, Wan~\cite{wang2025wan}+SR, as well as the native HunyuanVideo and Wan models. The supplementary results are illustrated in Fig.~\ref{fig:compare on 1k and 4k suppl}. As shown, both HunyuanVideo and Wan struggle to generate high-quality native 1080P and 4K videos: HunyuanVideo fails to follow the prompt and introduces significant noise at 4K resolution, while Wan produces videos that are overly smooth and lack detail. Although the super-resolution-based models are able to generate videos that are consistent with the prompts, their heavy reliance on super-resolution leads to outputs with reduced detail and texture. In contrast, our UltraGen model not only aligns closely with the given prompts but also achieves superior high-definition video generation quality.

\begin{figure*}[t]
    \centering
    \includegraphics[width=1.0\textwidth]{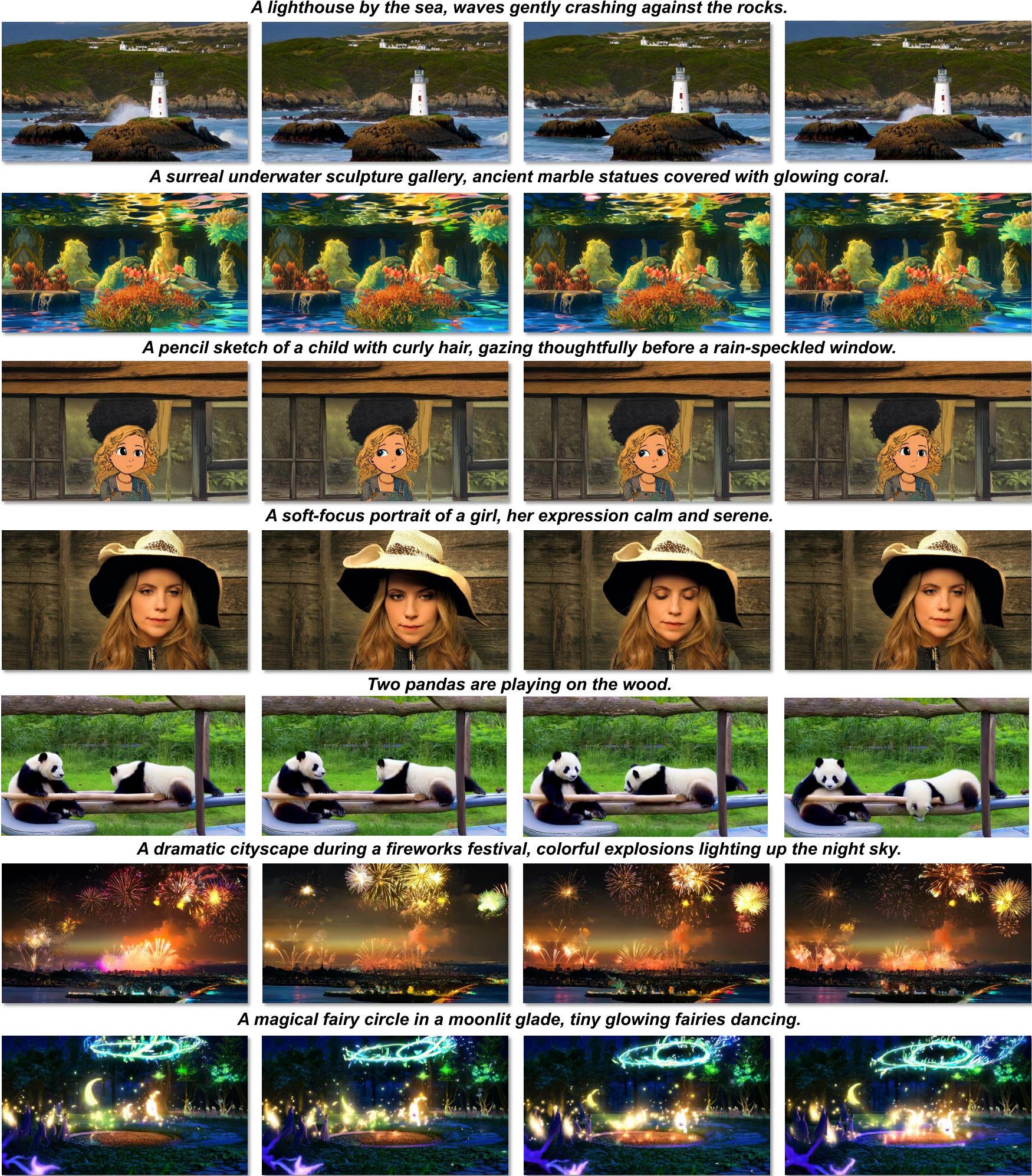}
    \caption{More \textbf{1080P} visualization results generated by our model.}
    \label{fig:more 1080P results}
\end{figure*}

\begin{figure*}[t]
    \centering
    \includegraphics[width=1.0\textwidth]{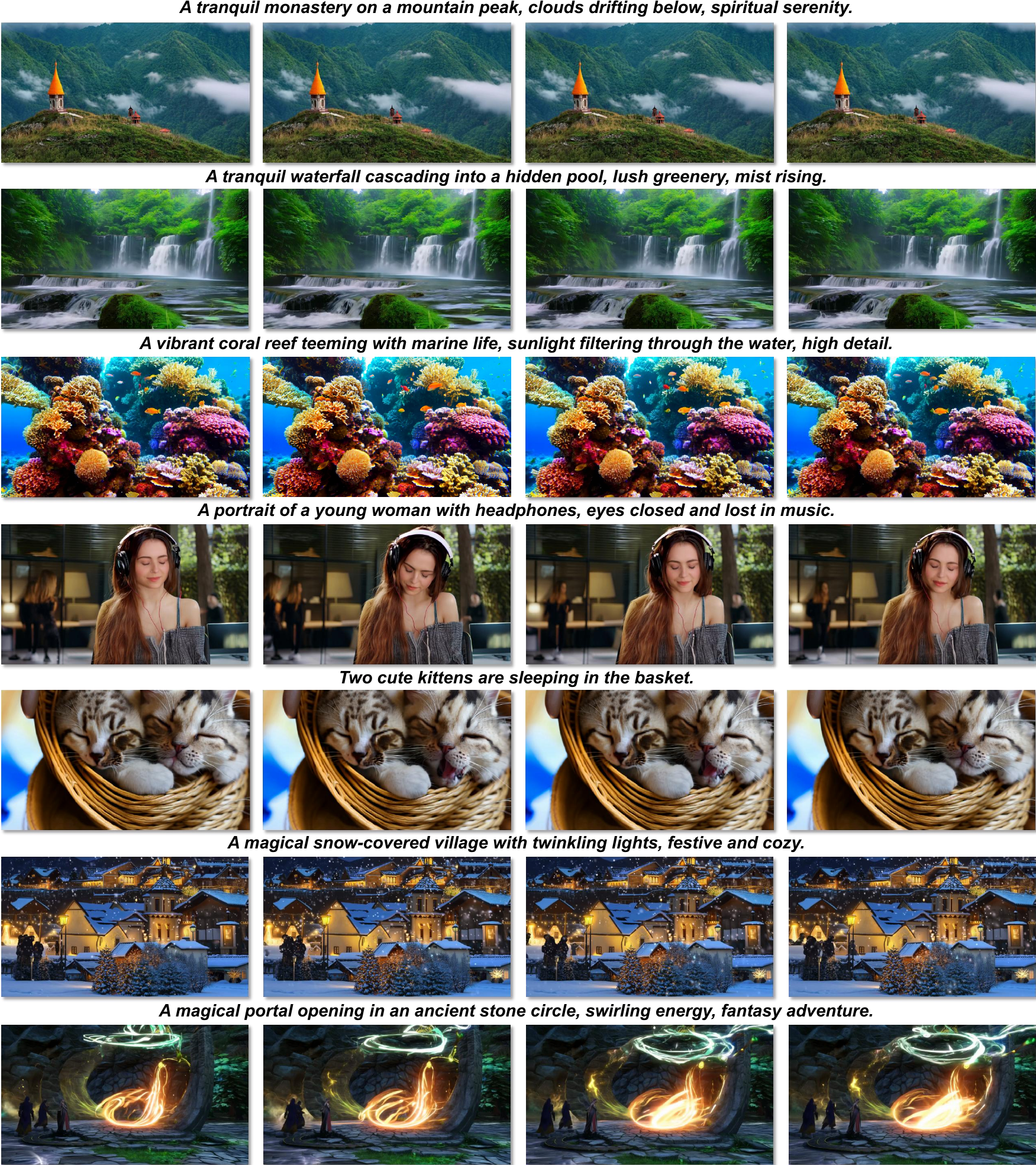}
    \caption{More \textbf{4K} visualization results generated by our model.}
    \label{fig:more 4k results}
\end{figure*}

\textbf{Additional 1080P and 4K Results.} To further demonstrate the effectiveness and robustness of our model, we present additional examples of generated 1080P videos in Fig.~\ref{fig:more 1080P results} and 4K videos in Fig.~\ref{fig:more 4k results}. As shown, our model consistently produces high-quality videos that faithfully correspond to a diverse range of text prompts. It should be noted that the 4K videos are limited to 29 frames due to GPU memory constraints, which may somewhat restrict their temporal dynamics. Nevertheless, the overall results are still very impressive.

\section{Limitations}
\label{sec:limitations}

While our model is capable of generating high-quality, high-resolution videos, it still inherits certain limitations from the underlying base model, which was originally designed for lower-resolution outputs. As a result, in particularly challenging scenarios—such as those involving rapid or large-scale motions—the model may occasionally encounter difficulties in accurately capturing complex motion dynamics, leading to minor artifacts or less natural motion. Addressing these challenges and further enhancing the model’s robustness in such demanding high-resolution settings will be an important focus of our future work.

% \begin{figure*}[t]
%     \centering
%     \includegraphics[width=1.0\textwidth]{figures/ablation.pdf}
%     \caption{Ablation study on the model without \textbf{\textit{(1)}} global attention; \textbf{\textit{(2)}} cross-window attention; \textbf{\textit{(3)}} hierarchical attention; and \textbf{\textit{(4)}} domain-aware LoRA. The model without global attention generates separate content within each local window. Models lacking cross-window or hierarchical attention suffer from temporal inconsistencies at window boundaries. Moreover, the model without domain-aware LoRA fails to produce videos of comparable quality to the full model.}
%     \label{fig:ablation study}
% \end{figure*}

% \input{../ReproducibilityChecklist/LaTeX/ReproducibilityChecklist.tex}

\end{document}